\documentclass[letterpaper]{article}
\usepackage[draft]{aaai25} 
\usepackage{times} 
\usepackage{helvet} 
\usepackage{courier} 
\usepackage[hyphens]{url} 
\usepackage{graphicx} 
\usepackage{graphicx} 
\usepackage{natbib} 
\usepackage{caption} 
\usepackage{tikz}
\usepackage{pgfplots}
\usepackage{amsmath}

\begin{document}
\title{Next Word Suggestion using Graph Neural Network}
\author{
    Abisha Thapa Magar\equalcontrib,
    Anup Shakya\equalcontrib
}
\affiliations {
    University of Memphis\\
    \{thpmagar, ashakya\}@memphis.edu
}
\maketitle

\begin{abstract}

Language Modeling is a prevalent task in Natural Language Processing. The currently existing most recent and most successful language models often tend to build a massive model with billions of parameters, feed in a tremendous amount of text data, and train with enormous computation resources which require millions of dollars. In this project, we aim to address an important sub-task in language modeling, i.e., context embedding. We propose an approach to exploit the Graph Convolution operation in GNNs to encode the context and use it in coalition with LSTMs to predict the next word given a local context of preceding words. We test this on the custom Wikipedia text corpus using a very limited amount of resources and show that this approach works fairly well to predict the next word.

\end{abstract}

\section{Introduction}

Language Model (LM) is a core component of Natural Language Processing. Deep Neural networks are widely used for building language models due to their strong representation power. Neural networks including the Feed-forward neural network and Recurrent neural network can automatically learn the features and continuous representation of text. The first feed-forward neural network used for Language Modeling~\cite{bengio2003} eliminated the curse of dimensionality by constructing a distributed representation for words, i.e., a vector called embedding. Eventually, recurrent neural networks~\cite{mikolov2010} entered the fray, and the research and development on DNNs for LM grew rapidly. Long Short-Term Memory (LSTM)~\cite{sundermeyer2012} eliminated the problem of diminishing gradients and improved learning long-term dependencies. More recently, attention mechanism introduced in the Transformers model~\cite{VaswaniSPUJGKP17} are very popular that has brought about a revolution in Language models. The large-scale models like BERT~\cite{bert2018}, GPT-3~\cite{NEURIPS2020_1457c0d6} and RoBERTa~\cite{roberta2019} are all based on the Transformers model.

Predicting the next word seems like a simple task, but  it is far from that as it requires a high level of understanding of the language. A naive approach to predicting the next word is to memorize the patterns in which different words co-occur and use frequency-based predictions. Clearly, this will not work in the real world. One of the simplest ways to comprehend language is to learn a deep representation of each word that encodes the context it appears. Representing the context in the text is a major challenge in this domain. In this project, we explore a new technique for embedding the context by representing text data as graph structures. Specifically, we construct a large graph from a corpus of text and implement Graph Neural Network (GNN) to learn context embeddings for each node. We exploit the natural property of the Graph Convolution~\cite{gcn2016} operation to accumulate the n-hop neighbors of each node and learn the node embeddings. We use these node embeddings and construct n-gram sequences for training the LSTM model. We test our approach on the custom Wikipedia corpus, which we create using the Wikipedia python package. The results show that this model trained on a small dataset with a minimum resource cannot compete with the state-of-the-art models, however, it can still yield outputs that are relevant to the corpus that it was trained on.

\section{Proposed Approach}
\subsection{Context Embeddings}
An embedding is a relatively low-dimensional dense vector representation of objects/entities. Word embeddings are an essential part of Language models. Each word is represented using a real-valued vector of tens or hundreds of dimensions. We can train the word embeddings such that they can reflect the semantics in language. For example, two words that have similar meanings can have similar word embeddings i.e., the distance between the two vectors is small. We can also train the vectors to represent the context in which they appear in the text. Here, we refer to context embeddings as the dense vector representations that encode the context of the words. In this project, we consider the context as two types: Global context and Local context. 

First, we create a graph structure from a given text corpus. We construct a graph $\mathcal{G} = (\mathcal{V}, \mathcal{E})$, where $\mathcal{E}$ is the set of nodes which represent the unique words in the text corpus and $\mathcal{V}$ is the set of vertices/links that represent the words that co-occur in the text. Let us take a toy example of a corpus below.

\noindent$\texttt{the weather is good.}$
$\texttt{the weather forecast is sunny.}$

\begin{figure}
    \centering
    \scalebox{0.25}{
        \includegraphics{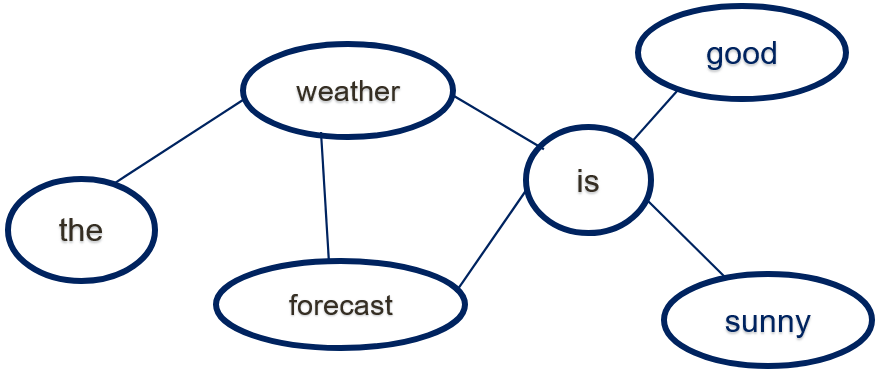}
    }
    \caption{Illustrating the graph structure constructed from the toy example. Each node represents the unique words in the corpus and links represent the co-occurrence of the words.}
    \label{fig:graph_structure}
\end{figure}

Figure \ref{fig:graph_structure} shows the constructed graph. The subsections below explain the idea of Global and Local context.

\subsubsection{Global Context}
Global context represents the general sense in which a word appears. For example in a news article about the current stock market, the global context for a word can be the set of words that either precede or follow it in the article. In the toy example above, as we can see in Figure \ref{fig:graph_structure},
the one-hop global context for the word \textit{\textbf{is}} are the words \textit{\textbf{weather}}, \textit{\textbf{forecast}}, \textit{\textbf{good}} and \textit{\textbf{sunny}}. Similarly, we can extract the two-hop neighbors and so on for every word in the corpus. We use the Graph Convolution operation to embed such neighbors into a dense vector representation.

\subsubsection{Local Context}
We can think of Local context as the neighborhood structure in a single sentence/paragraph. Let us take an example where a user is typing a search query in Google, and they have entered \texttt{which teams entered the knockout stage of the FIFA world}. Here, the local context for the word \textit{\textbf{world}} is all the words preceding it in the typed sentence. In such cases, the next word is more influenced by the shorter as well as longer-term dependencies. LSTMs work perfectly for such cases.

\subsection{Graph Convolution}

 In Convolutional Neural Networks (CNNs)~\cite{cnn2015}, convolution operation refers to multiplying the inputs with a set of weights that are commonly known as filters/kernels. The same set of weights is shared within a single layer. The filters act as a sliding window across the whole image and enable CNNs to learn features from neighboring pixels as illustrated in Figure 1. An image is a special case of a graph where a pixel is connected to its adjacent pixels. Similar to 2D convolution, one can perform graph convolution by taking the weighted average of a node's neighborhood information. So, the graph convolution operation is a generalized form of a 2D convolution operation.

 Graph convolutional operation can be summarized as:
 \begin{equation}
     \centering
     h^{i+1} = \sigma{(w^{i}h^{i}A)}
 \end{equation}
 where:
 
$h^{i+1}$: hidden layer representations of node at layer i+1

$w^{i}$: weights at layer i

$h^{i}$: hidden layer representations of node at layer i

$A$: adjacency matrix

$\sigma$: activation function

 In each layer of GCN, graph convolution captures a node's neighborhood information by one hop. Therefore, in the case of language modeling, graph convolution helps to preserve the context of a word.

\begin{figure}
    \centering
    \scalebox{0.23}{
        \includegraphics{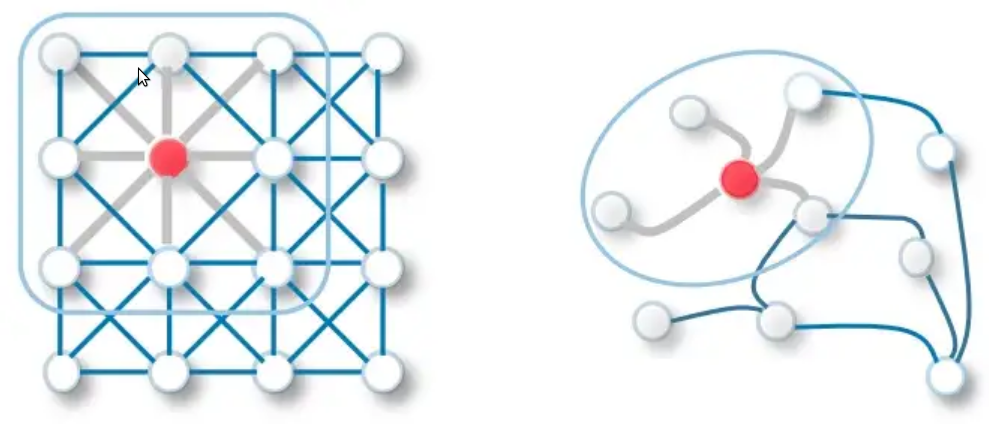}
    }
    \caption{Illustrating the convolution operation. Left image shows the 2D convolution in Convolutional Neural Networks. Right image shows the convolution in Graph Convolutional Networks [source:~\cite{gnn_survey2019}]}
    \label{fig:graph_conv}
\end{figure}

\begin{figure}
    \centering
    \scalebox{0.4}{
        \includegraphics{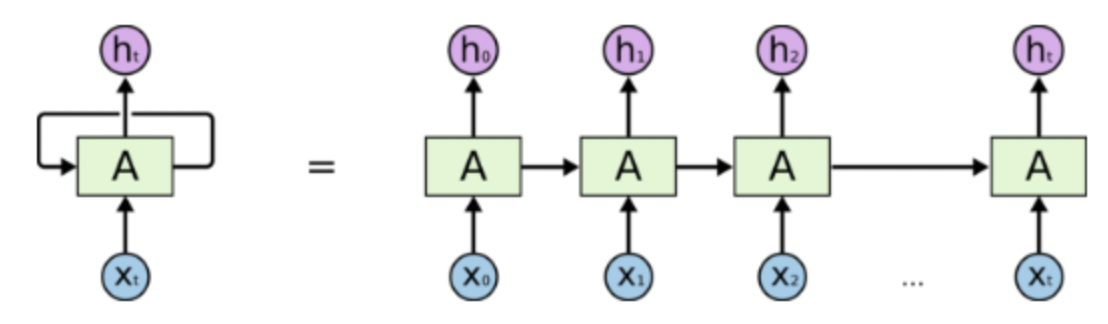}
    }
    \caption{Working of Recurrent Neural Network [Source:~\cite{rnn_article}]}
    \label{fig:rnn}
\end{figure}

\begin{figure*}
    \centering
    \scalebox{0.5}{
        \includegraphics{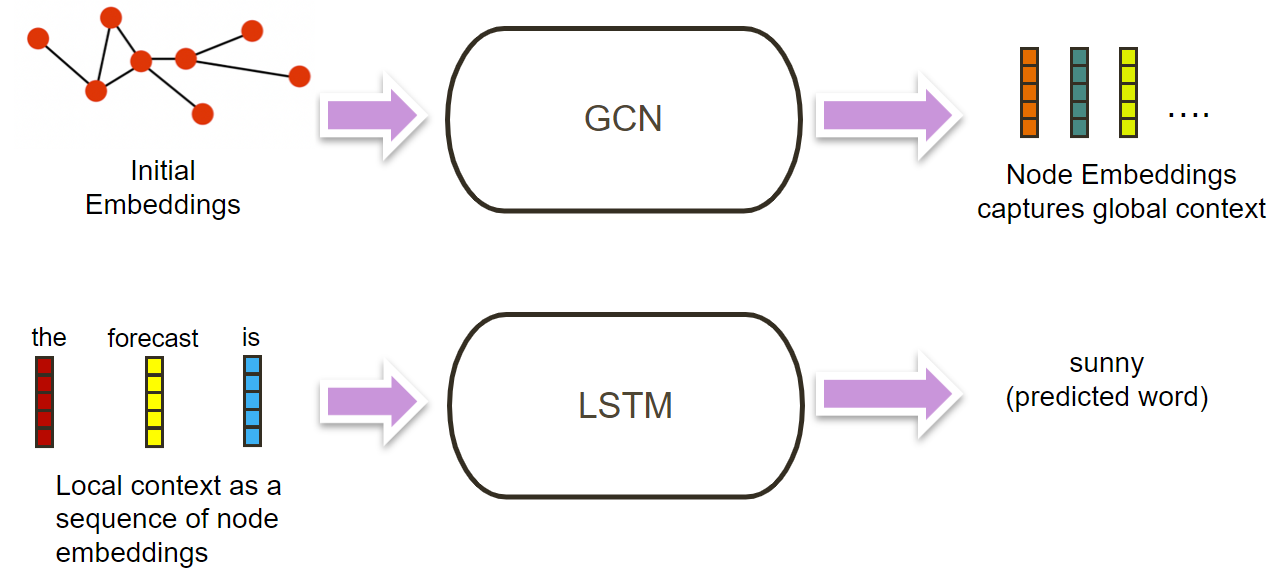}
    }
    \caption{Illustrating the overall architecture of the proposed approach for predicting the next word.}
    \label{fig:architecture}
\end{figure*}

\subsection{Recurrent Neural Networks(RNN)}
RNN is a feed-forward neural network that has an internal state and is recurrent in nature. It performs the same function for every input and the output of the current input depends on past computation (illustrated in Figure 3). RNNs use their internal state to process the sequence of inputs while in other neural networks, the sequence of inputs is independent of each other.

\subsubsection{Long Short Term Memory (LSTM) Networks}
LSTM networks are a special kind of RNN that is capable of learning long-term dependencies by remembering the information/sequence for a longer period of time. The vanishing gradient problem of RNN is resolved here. It uses back-propagation to train the model. LSTM consists of three gates: input gate, output gate, and forget gate.

Forget gate, $f_t$ decides which information to delete that is not important from the previous time step. The input gate, $i_t$ determines which information to let through based on its significance in its current time step. Output gate, $o_t$ allows the passed information to impact the output in the current time step. These gates can be realized using the following equations.

\begin{align}
    f_t &= \sigma(W_f\cdot[h_{t-1}, x_t]+b_f)\\
    i_t &= \sigma(W_i\cdot[h_{t-1}, x_t]+b_i)\\
    C_t &= \tanh(W_C\cdot[h_{t-1}, x_t]+b_C)\\ 
    o_t&= \sigma(W_o\cdot[h_{t-1}, x_t]+b_o)\\
    h_t &= o_t * \tanh(C_t)
\end{align}

where:

subscript [f, i, o, C] corresponds to forget, input, output, and cell state

$W_*$: Weights of the corresponding gate

$b_*$: bias of the corresponding gate

$x_t$: current input

$h_{t-1}$: previous hidden layer output

$h_t$: current hidden layer output

$\sigma$: activation function

\subsubsection{Many to One LSTM}
This LSTM takes a sequence of inputs and generates a single output. The previous inputs are considered as the context in LSTM and for our case, we have termed it as local context. To capture the local context efficiently, many-to-one LSTM can be trained on n-gram data.

\section{Experiments and Results}

\subsection{Dataset}

We use the Wikipedia corpus for our experiments. The entire Wikipedia corpus contains more than 1.9 billion words which takes large amount of resources for pre-processing and subsequent model training. Thus, we construct a custom Wikipedia dataset using the Wikipedia python package. We use the python package to extract different related articles using the in-built search query. For example, we use keywords like \textit{sports}, \textit{celebrity} to search for 1000 related articles and concatenate the articles to construct a single large corpus. Table \ref{tab:dataset_stat} shows the detailed statistics for the text corpus we use for the experiments.

Additionally, we pre-process the text corpus to obtain a clean train/test set. We perform the following steps for cleaning the data: (i) remove punctuation, (ii) remove extra white spaces, (iii) lowercase the data (iv) convert into word tokens. We construct graphs for each text corpora as explained in the toy example in Figure \ref{fig:graph_structure}.

\begin{table}[]
    \centering
    \caption{Dataset statistics for the custom Wikipedia corpus.}
    \label{tab:dataset_stat}
    \scalebox{0.8}{
    \begin{tabular}{|c|c|c|c|}
        \hline
        {\bf Keyword} & {\bf No. of words} &{\bf No. of unique words} & {\bf No. of Sentences}\\
        \hline\hline
        \textit{sports} & 65,485 & 8,365 & 3,563\\
        \hline
        \textit{celebrity} & 60,105  & 7,901 & 4,390\\
        \hline
        \textit{music} & 76,600  & 9,344 & 4,468\\
        \hline
    \end{tabular}
    }
\end{table}

\begin{table}[]
    \centering
    \caption{Parameters for the models.}
    \label{tab:param}
    \scalebox{0.8}{
    \begin{tabular}{|c|c|}
        \hline
        {\bf GCN} & {\bf LSTM}\\
        \hline\hline
        No. of GraphConv layers = 2 & Embedding dimension = 64\\
        
        Hidden Units Dimension = 64 & No. of hidden units = 200\\
        
        Adam Optimizer with lr = 0.005 & Adam optimizer with lr = 0.0001\\
        
        Train-test split of 80/20 & Train-test split 80/20\\
        
        No. of Epochs = 200 & No. of Epochs = 500\\
        
        & Batch Size = 100\\
        \hline
    \end{tabular}
    }
\end{table}


\subsection{Baseline Method}
The baseline method that we used as a control is the random embeddings plus LSTM combination. For this method, the word embeddings are simply initialized to random vectors and these embeddings are used in the input to the LSTM. This is a general method that people use for next-word prediction. We denote this method as Random Embedding (\texttt{RE}).

\subsection{Setup}
We used the GCN model to encode the global context of words and the LSTM model to encode the local context as explained in the sections above. Figure~\ref{fig:architecture} shows the overall setting of the approach we used. As shown in the figure, we train two models independently. For the GCN model, we initialize the embeddings of each node as random vectors of a fixed dimension and perform two-hop Graph convolution. We train the GCN model for the Link Prediction task. For the LSTM model, we train it with input sequences of different lengths. Specifically, we train it with 1-gram, 2-gram, 3-gram, 5-gram, and 10-gram sequences. Here, the sequential input for the LSTM model is constructed from the node embeddings learned from the GCN model. And we use post-padding with zeros to keep the sequence length fixed. We denote this method as Context Embedding (\texttt{CE}). Table~\ref{tab:param} illustrates the hyper-parameters for the two models. We train the models on a system with 64GB RAM, NVIDIA RTX Quadro 5000 GPU with 16GB memory, and 16-core Intel-i9 CPU.

\subsection{Results}
\begin{figure}
    \scalebox{0.98}{
    \begin{tikzpicture}
        \begin{axis} 
            [
            ybar,
            ybar=1pt, 
            bar width=7,
            ymin=20, 
            ymax=90, 
            ylabel={\textbf{Accuracy (\%)}}, 
            xlabel={\textbf{Text Corpora}},
            symbolic x coords={sports, celebrity, music},
            xtick = {sports, celebrity, music},
            legend pos=north west,
            legend style={nodes={scale=0.75, transform shape}},
            nodes near coords,
            nodes near coords align={vertical},
            nodes near coords style={font=\tiny, color=black}
            ]
    
            \addplot [green, fill] coordinates {(sports,52.6) (celebrity, 46.33) (music, 47.3)};
            \addplot [red, fill] 
            coordinates {(sports,26.43) (celebrity, 24.8) (music, 22.71)};
            \addplot [orange, fill] coordinates {(sports,59.56) (celebrity, 61.3) (music, 67.3)};
            \addplot [cyan, fill] 
            coordinates {(sports,34.56) (celebrity, 33.2) (music, 36.21)};
            
            \legend {RE-Train, RE-Test, CE-Train, CE-Test};
        
        \end{axis}
    \end{tikzpicture}}
    \caption{Bar chart showing the accuracy of next word prediction.}
    \label{fig:results}
\end{figure}
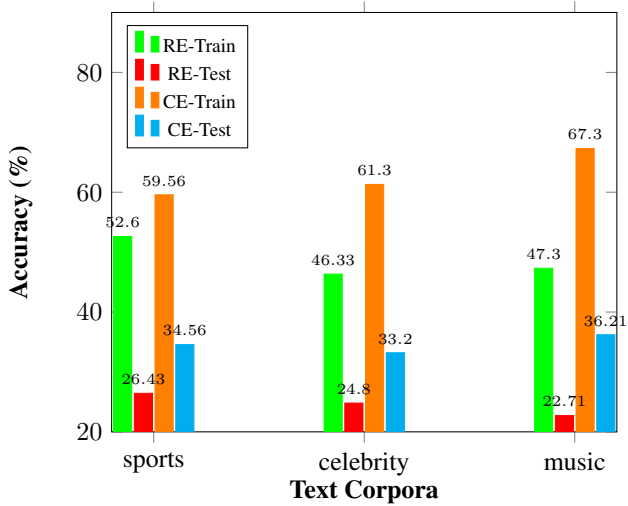

Figure~\ref{fig:results} shows the accuracy results for the next word prediction of the different text corpus. We can see that the model seems to learn the next word patterns in the training set. But, it does not seem to be able to generalize the patterns in the test set. This is the case for all three corpora. \texttt{CE} seems to perform better than \texttt{RE} which shows that vector embeddings are an essential part of modeling the patterns in the data. Overall, it seems that the models are still under-fitting the training data. Some of the predictions from the trained model are as follows:

\noindent\texttt{the new} $\rightarrow$ \texttt{bestselling}\\
\texttt{celebrities arrived at the} $\rightarrow$ \texttt{ownership}\\
\texttt{the guitar player was} $\rightarrow$ \texttt{spontaneously}\\
\texttt{the song browser} $\rightarrow$ \texttt{depends}\\
\texttt{rock music is rated very high in} $\rightarrow$ \texttt{opeth}\\

\section{Discussion}
We started this work as a research project and tried out novel ways to incorporate the context of words in a text into a vector representation and exploit the learned patterns to predict the next word given some local context. Due to various limitations of available resources to train large models with large text corpus, we ended up training the model with a fairly small dataset. And since language in general has many complex patterns that takes years even for humans to master, the models that we trained had very limited data and resource. This could be the primary reason for such a performance. Another reason could be the assumption that we made to use Graph Convolution to embed the global context. We will need to perform analysis like T-SNE plot on the learned node embeddings from GCN and run empirical tests to verify if the node embeddings encode the context or not. This will be a future task as we work on this research project. Another direction that we would like to explore is to try different types of Graph Neural Networks like GraphSAGE, GIN, GAT, and others.

\bibliography{main}
\end{document}